\pdfoutput=1

\documentclass[11pt]{article}

\usepackage{EMNLP2022}

\usepackage{times}
\usepackage{latexsym}
\usepackage{enumitem}
\usepackage{array,multirow}
\usepackage{float}
\usepackage[T1]{fontenc}

\usepackage[utf8]{inputenc}

\usepackage{microtype}

\usepackage{graphicx}
\usepackage{amsmath}
\usepackage{mathrsfs}
\usepackage{url}
\usepackage{booktabs,siunitx}
\newcommand{\argmax}{\operatornamewithlimits{argmax}}
\newcommand{\argmin}{\operatornamewithlimits{argmin}}
\usepackage{dsfont}
\newcommand{\mc}[3]{\multicolumn{#1}{#2}{#3}}

\usepackage{inconsolata}
\usepackage{amsfonts}

\title{Conversation Disentanglement with Bi-Level Contrastive Learning}

\author{Chengyu Huang   \\ National University of Singapore \\ e0376956@u.nus.edu
       \And
       Zheng Zhang \\ Tsinghua University \\zhangz.goal@gmail.com
        \AND
        Hao Fei \\ National University of Singapore \\ haofei37@nus.edu.sg
        \And
        Lizi Liao \\    Singapore Management University \\ lzliao@smu.edu.sg
        }

\begin{document}
\maketitle
\begin{abstract}
Conversation disentanglement aims to group utterances into detached sessions, which is a fundamental task in processing multi-party conversations. Existing methods have two main drawbacks. First, they overemphasize pairwise utterance relations but pay inadequate attention to the utterance-to-context relation modeling. Second, a huge amount of human annotated data is required for training, which is expensive to obtain in practice. To address these issues, we propose a general disentangle model based on bi-level contrastive learning. It brings closer utterances in the same session while encourages each utterance to be near its clustered session prototypes in the representation space. Unlike existing approaches, our disentangle model works in both supervised settings with labeled data and unsupervised settings when no such data is available. The proposed method achieves new state-of-the-art performance results on both settings across several public datasets.

\end{abstract}

\section{Introduction}
Multi-party conversations generally involve three or more speakers in a single dialogue, in which the speaker utterances are interleaved, and multiple topics may be discussed concurrently \citep{aoki2006}. This causes inconvenience for dialogue participant to digest the utterances and respond to a particular topic thread. Conversation disentanglement is the task of separating these entangled utterances into detached sessions, which is a prerequisite of many important downstream tasks such as dialogue information extraction \cite{ijcai2022p570,ijcai2022p571}, state tracking \cite{zhang2019neural,wu2022state}, response generation \cite{liao2018knowledge,liao2021mmconv,ye2022structured,ye2022reflecting}, and response ranking \citep{elsner2008,ryan2017}.
\begin{figure}
	\centering
	\vspace{+0.2cm}
	\includegraphics[scale=0.46]{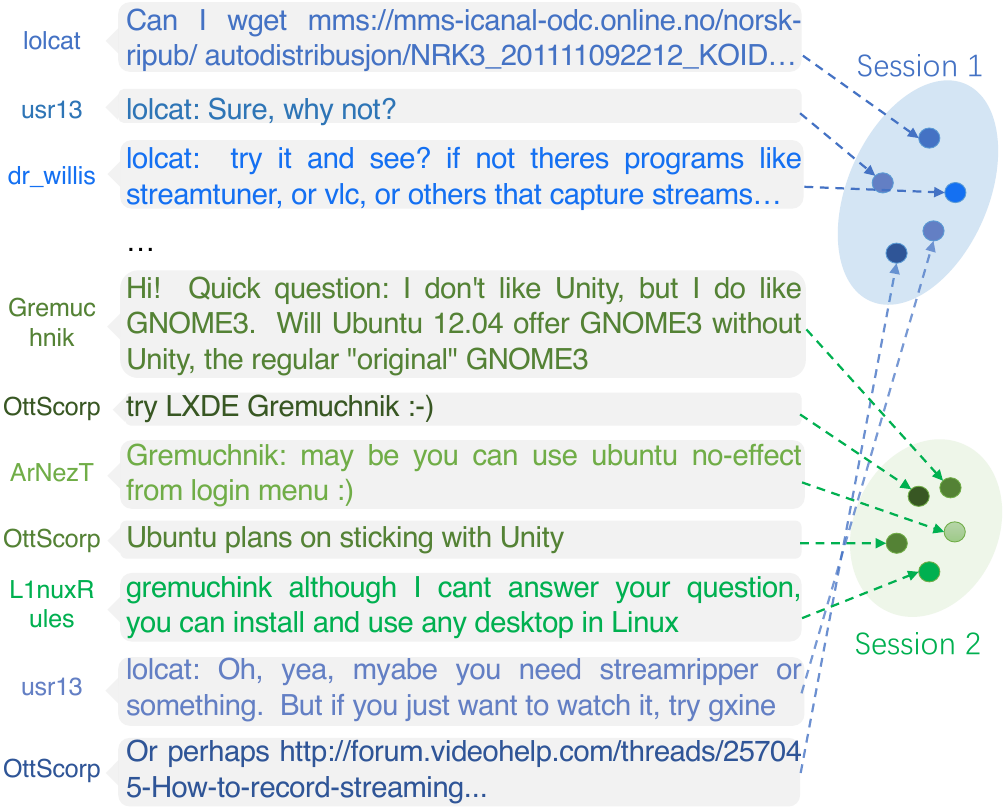}
	\vspace{-0.4cm}
	\caption{An example piece of conversation from the Ubuntu IRC corpus. There are distribution patterns in both utterance level and session level. }
	\vspace{-0.2cm}
	\label{example}
\end{figure}

There has been substantial work on the conversation disentanglement task. Most of them emphasize on the pairwise relation between utterances in a two-step manner. They predict the relationship between utterance pairs as the first step, followed by clustering utterances into sessions as the second. In the first step, early works \citep{elsner2008, elsner2010} utilize handmade features and discourse cues to predict whether two utterances belong to the same session or whether there is a reply-to relation. The recent development in deep learning inspires the use of neural network such as LSTM or CNN to learn abstract features of utterances in training ~\citep{mehri2017, jiang2018}. More recently, a number of methods show that BERT in combination with handcrafted features or heuristics remains a strong baseline \cite{li2020dialbert,zhu2021findings,ma2022structural}. In the second step, the most popular clustering methods use a greedy approach to group utterances by adding pairs \cite{wang2009context,zhu2020}. There are also some variations incorporating voting mechanism \cite{kummerfeld2019}, bipartite graph matching \cite{zhu2021findings} or additional tracking models \cite{wang2020response}.

An obvious drawback of such two-step approach is that the pairwise relation prediction might not capture enough contextual information as the connection between two utterances depends on the contexts in many cases \citep{ijcai2020-535}. Also, focusing on pairwise relations leads to a short-sighted local view. To mitigate this, there are methods trying to introduce additional conversation loss \cite{li2020dialbert,li2022conversation} or session classifier \cite{liu2021} to group utterances in the same session together. We also see methods leveraging relational graph convolution network \cite{ma2022structural} or masking mechanism in Transformers \cite{zhu2020}. More directly, end-to-end methods \cite{tan2019context,ijcai2020-535} capture the context information contained in detached sessions and calculate the matching degree between a session and an utterance. However, many of such methods are conducted in an online manner which only considers the preceding context. It may lead to biased session representations, introduce noisy utterances to sessions and consequently accumulate errors.

Meanwhile, most of these methods rely heavily upon human-annotated session labels or reply-to relations, which are expensive to obtain in practice. Although there have been a few attempts to tackle this issue, a more general framework that can handle both supervised and unsupervised learning is yet to come. For example, \citet{liu2021} design a deep co-training scheme with message-pair classifier and session classifier. However, various data augmentation procedures based on heuristics are required for good performance. \citet{chi2021zero} propose a zero-shot disentanglement solution based on a related response selection task. Still, it relies on a closely related dataset that comes from the same Ubuntu IRC source inside DSTC8.

Recently, contrastive learning \cite{hadsell2006dimensionality} has brought prosperity to numbers of machine learning tasks by introducing unsupervised representation learning. Substantial performance gains have been reported in computer vision \cite{he2020momentum,chen2020simple} and NLP works \cite{yan2021consert,gao2021simcse}. They believe that good representation should be able to identify semantically close neighbors while distinguishing from non-neighbors. Intuitively, in multi-party conversation, utterances in the same session should semantically resemble each other while be far apart from utterances in other sessions. Instead of hand-crafted features such as speaker, mention and time difference \textit{etc}, it provides another option for automatically learn discriminative representations.

In this work, we design a Bi-level Contrastive Learning scheme (Bi-CL) to learn discriminative representations of tangled multi-party dialogue utterances. It not only learns utterance level differences across sessions, but more importantly, it encodes session level structures  discovered by clustering into the learned embedding space. Specifically, we introduce session prototypes to represent each session for capturing global dialogue structure and encourage each utterance to be closer to their assigned prototypes. Since the prototypes can be estimated via performing clustering on the utterance representations, it also supports unsupervised conversation disentanglement under an Expectation-Maximization framework. We evaluate the proposed model under both supervised and unsupervised settings across several public datasets. It achieves new state-of-the-art on both.

The contribution is summarized as follows:
\begin{itemize}[itemsep=2pt,topsep=1pt,parsep=0pt]
	\item We design a bi-level contrastive learning scheme to learn better utterance level and session level representations for disentanglement.
	\item We delve into the conversation nature to harvest evidence which supports our model to disentangle dialogues without any supervision.
	\item Experiments show that the proposed Bi-CL model significantly outperforms several state-of-the-art models both on the supervised and unsupervised settings across datasets.
\end{itemize}
\section{Related Work}
\subsection{Conversation Disentanglement}
Previous methods on conversation disentanglement are mostly performed in a supervised fashion, which can be coarsely organized into two lines: (1) two-step methods which first obtain the pairwise relations among utterances and then disentangle them with a clustering algorithm; and (2) end-to-end approaches which directly assign utterances into different sessions.

The majority of efforts follow the two-step pipeline. Great attention has been devoted to the first step. Early works rely heavily on handcrafted features to represent the utterances for pairwise relation prediction. For example, \citet{elsner2008, elsner2010} used the speaker, time, mentions, shared word count \textit{etc.} to train a linear classifier for utterance pair coherence. More recent works utilized neural networks to train classifiers. For instance, \citet{mehri2017} and \citet{guo2018} leveraged LSTM to predict either the same-session or reply-to probabilities between utterances, while \citet{jiang2018} combined the output of a hierarchical CNN on utterances with other features to capture the interactions. More recently, \citet{gu2020} and \citet{li2020dialbert} used BERT to learn the similarity score in a fixed length context window. For the second step, there has also been progress in exploring an optimal clustering algorithm. Greedy decoding has been a popular choice \citep{elsner2010, jiang2018}. There are also works that train a separate classifier to assign utterance to a thread \citep{mehri2017} or design advanced algorithms like bipartite graph matching \cite{zhu2021findings}.

On the downside, the pairwise relations, which are predicted typically without considering enough session context, are local and may not reflect how utterances interact in reality. Hence, the clustering step may be undermined subsequently. This motivates end-to-end solutions that aim at assigning the target utterance in each time step with respect to the existing threads or preceding utterances \cite{ijcai2020-535}. 
Similarly, \citet{yu2020} used attention to capture utterance interactions and gradually assign each utterance to its replied-to parent with a pointer module. However, such online manner not only limits the scope of session context but also leads to error accumulation. 

There are also studies that work in an unsupervised fashion to avoid the reliance on human-annotation. For example, \citet{liu2021} designed both message-pair classifier and session classifier to form a co-training algorithm. \citet{chi2021zero} proposed to train a closely-related response selection model for zero-shot disentanglement. The former needs pseudo labeled data to warm-up the training, while the latter gains from training data of the same source. More importantly, a general framework that can handle both supervised and supervised learning is yet to come. In our work, we target at building such a flexible model.

\subsection{Contrastive Learning}
Contrastive learning learns effective representation by pulling semantically close neighbors together and pushing apart non-neighbors \cite{hadsell2006dimensionality}. Recent advances are largely driven by instance discrimination tasks. For example, in the field of computer vision, such methods consist of two key components: image transformation and contrastive loss. The former aims to generate multiple representations about the same image, by data augmentation \cite{ye2019unsupervised,chen2020simple}, patch perturbation \cite{misra2020self}, or using momentum features \cite{he2020momentum}. While the latter aims to bring closer samples from the same instance and separate samples from different instances. In the field of natural language processing, contrastive learning has also been widely applied, such as for language model pre-trainining \cite{yan2021consert,gao2021simcse}.

Despite their improved performance, these instance discrimination methods share a common weakness: the representation is not encouraged to encode the global semantic structure of data \cite{caron2020unsupervised}. This is because it treats two samples as a negative pair as long as they are from different instances, regardless of their semantic similarity \cite{li2020prototypical}. Hence, there are methods which simultaneously conduct contrastive learning at both the instance- and cluster-level \cite{li2021contrastive,shen2021you}. Likewise, we emphasize leveraging bi-level contrastive objects to learn better utterance level and session level representations.

\section{Method}
The definition of the conversation disentanglement task and details of our model are sequentially presented in this section. Starting from the supervised setting for a clear view, we gradually extend to the unsupervised setting.
\begin{figure*}
	\centering
	\includegraphics[scale=0.7]{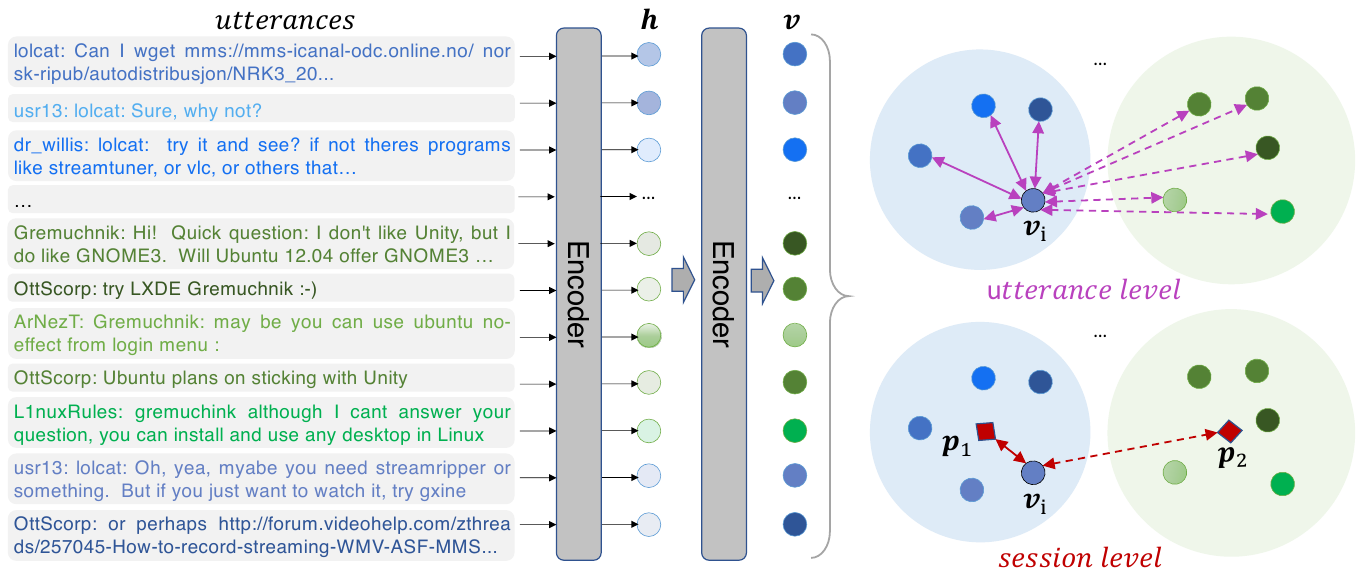}
	\caption{Overview of the proposed Bi-CL framework. It incorporates utterance level contrastive loss to discriminate utterances, and uses session level contrastive loss to encourage them flocking around session centers.}
	\label{model}
\end{figure*}

\subsection{Task Formulation}
Given a multi-party conversation history with $n$ utterances $U = \{u_1, u_2, ..., u_n\}$ in chronological order, our goal is to disentangle them into detached sessions $S = \{s_1, s_2, ..., s_k\}$, where each $s_i$ is a non empty subset of $U$, and $S$ is a partition of $U$. 
Each utterance includes an identity of speaker and a message sent by this user. 

The task has been popularly formulated as a reply-to relation identification problem to find the parent utterance for every $u_i \in U$. It has also been modeled as sequentially assigning each $u_i$ to already detached sessions in $S$ or create a new session for $S$. Here, instead of separating local pair and global cluster modeling, we opt for learning more discriminative representations for utterances to push them into different sessions.

\subsection{Utterance Encoder}
The utterance encoder aims to capture the semantics of a given utterance and its connection to surrounding context. Similar to \cite{ijcai2020-535}, we leverage a hierarchical Bi-LSTM structure similar to \cite{serban2017hierarchical} as illustrated in Figure \ref{model}.

For the utterance-level  encoder, given each utterance $u_i$, we tokenize it into tokens $\{t_1, t_2, ..., t_{|u_i|}\}$ and take the Glove embeddings \citep{glove2014}. We input these into a bidirectional LSTM and then use a linear transformation with non-linear activation to get the hidden states:
\begin{align*}
	\langle \textbf{h}_1, ..., \textbf{h}_{|u_i|}\rangle &= \delta(\textbf{W}_1\cdot BiLSTM(\langle t_1, ..., t_{|u_i|}\rangle)),\\
	\textbf{u}_i &= SelfAttention(\textbf{h}_1, \cdots, \textbf{h}_{|u_i|}),
\end{align*}
where $\textbf{W}_1$ is the weight matrix that merges the two direction embeddings of each token, and we use ReLU as the activation $\delta$. We omit the bias term for space limitation. The self-attention mechanism \citep{lin2017} is adopted to obtain utterance vectors that represent the overall semantics.

For the context-level encoder, we leverage another bidirectional LSTM to allow utterances to interact with their surroundings and acquire contextual information. Hence, we feed in the local utterance embedding sequence $\langle \textbf{u}_1, ...,  \textbf{u}_n\rangle$ and obtain the contextual utterance representations $\langle \textbf{h}'_1, ..., \textbf{h}'_n\rangle $. It naturally captures information in the utterance itself, in its surrounding utterances and the relative temporal sequence implicitly as:
\begin{equation*}
	\langle \textbf{h}'_1, ..., \textbf{h}'_n\rangle = BiLSTM(\langle \textbf{u}_1, ...,  \textbf{u}_n\rangle) .
\end{equation*}

To further utilize the speaker and mention information of each utterance, we simply concatenate each $\textbf{h}'_i$ with a padded, multi-hot mention vector $\textbf{m}_i \in R^{50}$ where the $j$-th dimension is 1 if the speaker of $u_j$ is the same as that of $u_i$ or mentioned in $u_i$. This will give the final utterance representations $\langle \textbf{v}_1, ..., \textbf{v}_n\rangle$.

\subsection{Bi-Level Losses}
With encoder network ready at hand, the key is to introduce good objectives for back-propagating the right learning signals. When we have session labels for training data in the supervised learning setting, we aim to train the model so that ideally, (a) utterances in the same ground truth session will be embedded closer while utterances in different sessions will be pulled away; and (b) utterances in each session should be near its session center, or say, prototype. Correspondingly, we introduce utterance-level contrastive loss and session-level contrastive loss to encourage these for learning.

\subsubsection{Utterance-level Contrastive Loss}
Inspired from the contrastive learning scheme ~\citep{khosla2020supervised} under supervised setting, we contrast an utterance with other utterances in same or different sessions to capture the local structure. 
Suppose the training dataset $\mathcal{U}$ contains $|\mathcal{U}|$ utterances in total and $y(i)$ denotes the ground truth session assignment of $u_i$, we define the utterance-level contrastive loss as:
\begin{align*}
		L_{u} = \sum_{i=1}^{|\mathcal{U}|} \frac{-1}{|\mathcal{Y}(i)|}  \sum_{j\in \mathcal{Y}(i)} log \frac{exp(\textbf{v}_i\cdot \textbf{v}_j/\tau_1)}{\sum\limits_{l\in\mathcal{N}(i, j)} exp(\textbf{v}_i\cdot \textbf{v}_l/\tau_1)},
\end{align*}
where $\tau_1$ is the temperature hyper parameter, $\mathcal{Y}(i)$ contains all the positive utterances that have the same session assignment with $u_i$, and $\mathcal{N}(i, j)$ contains the set of negative utterances that have different session assignment with $u_i$, combined with the current positive utterance $u_j$. Mathematically, we have $\mathcal{Y}(i) \equiv \{j\in \mathcal{U}: y(j) = y(i), j\ne i\}$, and $\mathcal{N}(i, j)  \equiv  \{l\in \mathcal{U}: y(l) \ne y(i)\} \cup \{j\}$. Ideally, we could use all negative samples as many papers have shown increased performance with increasing number of negatives \cite{he2020momentum,henaff2020data}, we set a relatively large number for balancing our computation efficiency.

\subsubsection{Session-level Contrastive Loss}
In session level, we introduce prototypes to represent each session, and minimize the distance from each utterance to its session prototype while maximize the distances from the utterance to other session prototypes. This incorporates global dialogue semantic structure into the resulting representations. When session labels are available in the supervised setting, suppose $s_i = \{u_{1}, u_{2}, ..., u_{q} \}$,  we directly define the prototype $p$ for session $s_i$:
\begin{equation*}
	\textbf{p} = \frac{1}{|q|}\sum_{j=1}^{q} \textbf{v}_{j} ~.
\end{equation*}

Therefore, for each conversation $U$ in the training set, we define the session-level contrastive loss:
\begin{align*}
		L_{s} = -\sum_{i=1}^{|U|} \log \frac{exp(\textbf{v}_i\cdot \textbf{p}_i/\tau_2)}{\sum\limits_{\textbf{p}_l} exp(\textbf{v}_i\cdot \textbf{p}_l/\tau_2)},
\end{align*}
where $\textbf{p}_i$ is the ground truth session prototype for $u_i$ and $\tau_2$ is the temperature hyper parameter.

\subsection{Disentangle Sessions}
Besides guiding the learning process with bi-level contrastive objects, our disentanglement task naturally involves the session assignment goal. Therefore, the foremost issue is to decide how many sessions the conversation contains. With supervised data, we train a light-weight network to predict $K$ for each conversation. We leverage a two layer feed-forward network enriched with non-linearity. It takes as input the dialogue utterances as well as meta information such as number of speakers $n_s$ and turn number $n$. The output logits indicate a distribution of the possible \textit{K} values.
\begin{align*}
	\textbf{d}_{U} &= SelfAttention(\textbf{v}_1,  \cdots, \textbf{v}_n),\\
	\textbf{q} &= \delta(\textbf{W}_2\cdot\delta(\textbf{W}_3\cdot[\textbf{d}_{U};~n_s;~n])),\\
	P(K=k|U) &= \frac{\exp(\textbf{q}_{k})}{\sum_{l=1}^{M} \exp(\textbf{q}_{l})},
\end{align*}
where $M$ is the global maximum session number, and $\textbf{q} \in \mathbb{R}^{M}$. We train the network parameters including $\textbf{W}_2$, $\textbf{W}_3$ via the $K$ prediction loss:
\begin{align*}
	L_k = -\sum_{U\in\mathcal{U}} \log(P(K = \hat{k}|U)),
\end{align*}
where $\hat{k}$ is the ground truth \textit{K}  for conversation $U$. In inference, we select the most likely value of \textit{K} for the K-Means algorithm and constrain $K <= n$.

During training, we also perform K-Means to cluster utterances to mitigate the gap between training and inference. Suppose we obtain a partition $S' = \{s'_1, s'_2, ..., s'_{\hat{k}}\}$ for the conversation $U$ by K-Means, we compute the cluster centroids $\{\textbf{c}'_1, \textbf{c}'_2, ..., \textbf{c}'_{\hat{k}}\}$ by averaging the embeddings of cluster members. We then run Hungarian Algorithm \cite{kuhn1955hungarian} to match clusters with sessions, hence align the calculated prototypes with these centroids, \textit{e.g.} $\textbf{p}_i$ to $\textbf{c}'_i$. We further introduce a centroid matching loss:
\begin{align*}
	L_m = \sum_{U\in\mathcal{U}} \frac{1}{\hat{k}}\sum_{i=1}^{\hat{k}} \|\textbf{p}_i - \textbf{c}'_i\|,
\end{align*}
which ensures that utterance embeddings are clustered according to their ground truth sessions.

To sum up, the final objective for supervised training is as below:
\begin{align}
	L_{supervised} = L_u + \alpha L_s + \beta L_k + \gamma L_m,
\end{align}
where $\alpha, \beta, \gamma$ are hyper-parameters to adjust the contribution of different factors. 

\subsection{Unsupervised Extension}
In the unsupervised setting, we mainly update the bi-level losses $L_{u}$ and $L_{s}$ for representation learning while omit the $L_k$ and $L_m$ losses. In the session level, since we do not know the session labels anymore, we directly estimate the session assignment by clustering utterance embeddings, and then maximize the data log-likelihood. Inspired from \cite{li2020prototypical}, we perform the two steps iteratively to form an Expectation-Maximization framework. The following shows our objective under the framework. More derivation details can be found in Appendix \ref{sec:appendix}.

In a specific iteration, suppose we obtain cluster results as $\{c_1, ..., c_m\}$ by running K-Means on conversation $U$, maximizing log-likelihood estimation corresponds to finding the utterance encoder network parameters that minimizes the loss:
\begin{align*}
	-\sum_{i=1}^{|U|} \log \frac{exp(\textbf{v}_i\cdot \textbf{c}_i/\phi_i)}{\sum_{l=1}^m exp(\textbf{v}_i\cdot \textbf{c}_l/\phi_l)},
\end{align*}
where $\phi$ denotes the concentration level of the feature distribution around a cluster centroid $c$. It encourages utterances to flock around the centroids. 

In practice, we cluster the utterances $M$ times with different number of clusters $K = \{k_m\}^M_{m=1}$, to achieve a more robust probability estimation of prototypes. Hence the updated session level loss is calculated as:
\begin{align*}
	L'_s = -\frac{1}{M} \sum_{i=1}^{|U|} \sum_{m=1}^{M}\log \frac{exp(\textbf{v}_i\cdot \textbf{c}_i/\phi_i)}{\sum_{l=1}^{k_m} exp(\textbf{v}_i\cdot \textbf{c}_l/\phi_l)},
\end{align*}
since the number of utterances in conversation $U$ is limited, we set $\phi$ to a small constant $\tau'_2$.

In the utterance-level, we make use of heuristics to construct positive and negative samples for contrastive learning. The assumption is that one speaker mostly participates in only one session \footnote{Only 20\% of speakers will join multiple sessions on the Ubuntu IRC dataset.}, and utterances in different conversations are naturally in different sessions. Suppose the speaker of $u_i$ is $s(i)$ in the conversation $U_i$, we update the utterance level contrastive loss as below:
\begin{align*}
	L'_{u} = \sum_{i=1}^{|\mathcal{U}|} \frac{-1}{|\mathcal{Y}'(i)|}  \sum_{j\in \mathcal{Y}'(i)} log \frac{exp(\textbf{v}_i\cdot \textbf{v}_j/\tau'_1)}{\sum\limits_{l\in\mathcal{N}'(i, j)} exp(\textbf{v}_i\cdot \textbf{v}_l/\tau'_1)},
\end{align*}
where $\mathcal{Y}'(i) \equiv \{j\in U_i: s(j) = s(i), j\ne i\}$, and $\mathcal{N}'(i, j)  \equiv  \{l\in \mathcal{U}/U_i\} \cup \{j\}$. To sum up, the final objective for unsupervised training is as below:
\begin{align}
	L_{unsupervised} = L'_u + \eta L'_s ,
\end{align}
where $\eta$ is a hyper-parameter to adjust the contribution of different factors. 

After the representation learning, we may use various methods to decide the session number $k$ for each conversation, such as the Elbow algorithm \cite{thorndike1953belongs}, or Silhouette algorithm \cite{rousseeuw1987silhouettes}. Empirically, we find the Elbow algorithm works slightly better. Based on the predicted $K$, we simply run the K-Means clustering to obtain the session assignments. 

\section{Experiments}
\subsection{Dataset}
We train and evaluate our models on two large-scale annotated datasets. The first dataset is the Ubuntu IRC dataset \citep{kummerfeld2019}, which consists of 153/10/10 intermingled dialogues in the train/validation/test set. Each dialogue is extracted from the Ubuntu IRC technical support channel and has a length of 250 or 500. Following \cite{ijcai2020-535}, we cut each dialogue into dialogue segments of length 50, reorder the ground truth session labels, and get 1,737/134/104 dialogues in the train/validation/test split. The maximum session number is 14 for the Ubuntu IRC dataset. The second dataset is the Movie Dialogue dataset \citep{ijcai2020-535}. The dialogues are generated by extracting sessions from 869 movie scripts and manually intermingling the sessions. There are 29,669/2,036/2,010 dialogues train/validation/test split. The maximum session number is 6.

\subsection{Training Details}
We initialize the word embeddings with 300-dimensional Glove vectors \citep{glove2014} and set the hidden state size of BiLSTM to be 300. The utterance embedding size after the co-attention layer will also be 300. The maximum length of an utterance after tokenization is set to 50. In supervised training, the hyperparameter $\alpha$ and $\beta$ that controls the weights are configured as 0.4 empirically, while $\gamma$ is set to 0.2. We adopt a batch size of 16 and use Adam Optimizer \citep{adam2014} with an initial learning rate of 5e-5. We run ten epochs until convergence. In unsupervised training, the hyper parameters will be the same and $\eta$ is set to 0.4. While certain hyper-parameters such as Glove embedding size are set according to the default practice of previous works, other hyper-parameters such as batch size and maximum sequence length are determined empirically. In particular, the weight parameters $\alpha$, $\beta$, $\gamma$, and $\eta$ are tuned with grid search.

\begin{table*}
	\centering
	\small
	\setlength{\tabcolsep}{13pt}
	\renewcommand*{\arraystretch}{1.1}
	\begin{tabular*}{0.96\textwidth}{clccc|ccc}
		\toprule
		&& \mc{3}{c}{Ubuntu IRC} & \mc{3}{c}{Movie Dialogue}\\
		\cmidrule{3-5} \cmidrule{6-8}
		&& {$NMI$} & {$ARI$} & {$Shen-F$} & {$NMI$} & {$ARI$} & {$Shen-F$} \\
		\midrule
		\parbox[c]{1mm}{\multirow{10}{*}{\rotatebox[origin=c]{90}{Supervised}}} & \textit{Weighted SP*} &   0.253 &   0.026 &   0.333 &   0.184 &    0.041 &    0.523\\
		& \textit{CISIR*} &  0.466 &  0.034 &   0.408 &  0.205 &  0.065 &   0.538\\
		& \textit{BERT*} & 0.546 & 0.082 & 0.439 &0.256  & 0.110 &0.569 \\
		& \textit{Transition} & 0.611 & 0.198 & 0.538 & 0.329 & 0.248 & 0.650\\
		& \textit{DialBERT} & 0.675 & 0.245 & 0.605 & 0.362 & 0.180 & 0.597\\
		& ~~~~\textit{+ cov} & \textbf{0.696} & 0.275 & 0.615 & 0.328 & 0.180 & 0.608\\
		& ~~~~\textit{+ feature} & 0.671 & 0.216 & 0.586 & - & -&-\\
		& ~~~~\textit{+ future context} & 0.671 & 0.226 & 0.591 & 0.358 & 0.174 & 0.587\\
		& \textit{StructBERT} & 0.678 & 0.371 & 0.675 & 0.446 & 0.327 & 0.695\\
		& ~~~~\textit{w/max-pooling} & 0.677 & \textbf{0.379} & 0.681 & 0.448 & 0.327 & 0.695\\
		& \textit{Bi-CL~(Ours)} &   0.624 &   0.360 &   \textbf{0.707} &   \textbf{0.575} &   \textbf{0.382} &    \textbf{0.747}\\
		\midrule
		\parbox[c]{1mm}{\multirow{3}{*}{\rotatebox[origin=c]{90}{Unsupervised~}}} & \textit{Co-Training} & 0.540 & 0.182 & 0.456 & 0.290 & 0.217 & 0.592\\
		&~~~~\textit{- pseudo data} & 0.531 & 0.168 & 0.409 & 0.279 & 0.201 & 0.576\\
		&\textit{Zeroshot} & 0.597  & 0.212 &  0.578 & - & - & -\\
		&~~~~\textit{+ augment data} & \textbf{0.639}  & 0.292 & 0.642  & - & - & -\\
		&\textit{Bi-CL~(Ours)}&   0.607 &   \textbf{0.345} &   \textbf{0.704} &   \textbf{0.581} &   \textbf{0.366} &    0.689\\
		&~~~~\textit{w/Silhouette} & 0.606  & 0.343 & 0.704  & 0.562 & 0.352 & \textbf{0.698}\\
		\bottomrule
	\end{tabular*}
	\caption{\label{result} Results on the Ubuntu IRC Dataset and the Movie Dialogue Dataset. * indicates that the statistics are taken from \citep{ijcai2020-535}. Note the results of DialBERT + feature on the Movie Dialogue Dataset is not available since the dataset does not provide the corresponding features.}
\end{table*}

\subsection{Metrics}
We adopted three popular metrics to evaluate the disentanglement result: Normalized Mutual Information (\textit{NMI}), Adjusted Random Index (\textit{ARI}) \citep{ari1985}, and Shen-F score (\textit{Shen-F}) \citep{shen2006}. Both \textit{NMI} and \textit{ARI} measures the similarity between the ground truth clusters and the predicted clusters for each conversation and a higher value indicates higher degree of matching. The difference is that \textit{ARI} is based on counting pairwise links between utterances that exist in both ground truth and predictions, while \textit{NMI} is more about the cluster level since it uses entropy conditioned on clusters. \textit{Shen-}F is a F-1 score to measure how well utterances in the same ground truth cluster are grouped in the predicted clusters, and a higher value indicates higher cluster quality.

\subsection{Baseline Models}
We evaluate on both supervised and unsupervised settings. The baselines include both the traditional two-stage based and end-to-end approaches.

 \textbf{Supervised Baselines:} The majority of methods need supervision. \textit{Weighted SP} \cite{shen2006} adopts a single pass greedy decoding to add and cluster utterances sequentially based on normalized TF-IDF vectors.
\textit{CISIR} \cite{jiang2018} uses Hierarchical CNN to encode utterances and compute score of pairs. \textit{Transition} \cite{ijcai2020-535} is an end-to-end online approach where each utterance is encoded and compared with the existing session states to determine assignments. \textit{DialBERT} \cite{li2020dialbert} gains from hierarchical Pre-Trained model for better performance.
 \textit{StructBERT} \cite{ma2022structural} emphasizes structural characteristics in modeling and is the current state-of-the-art.

\textbf{Unsupervised Baselines:}  When no labeled data is available, \textit{Co-Training} \cite{liu2021} leverages a message-pair classifier and session classifier to build up a co-training scheme. \textit{Zeroshot} \cite{chi2021zero} learns from a closely related response selection task.

\subsection{Main Results}
We report the main results for all compared methods in Table \ref{result}. Generally speaking, the proposed \textit{Bi-CL} method performs better than all the other baselines on both the Ubuntu IRC and Movie Dialogue datasets in most evaluation metrics. Note that some of these baselines are based on large-scale pre-trained language model BERT which has shown superior performance on various NLP tasks, our model is only based on the relatively  light-weight bidirectional LSTM model. This situation, in some sense, signals the effectiveness of our bi-level contrastive learning design.

\begin{table*}[!htp]
	\centering
	\small
	\setlength{\tabcolsep}{13pt}
	\renewcommand*{\arraystretch}{1.1}
	\begin{tabular*}{0.96\textwidth}{clccc|ccc}
		\toprule
		&& \mc{3}{c}{Ubuntu IRC} & \mc{3}{c}{Movie Dialogue}\\
		\cmidrule{3-5} \cmidrule{6-8}
		&& {$NMI$} & {$ARI$} & {$Shen-F$} & {$NMI$} & {$ARI$} & {$Shen-F$} \\
		\midrule
		\parbox[c]{1mm}{\multirow{6}{*}{\rotatebox[origin=c]{90}{Supervised}}}
		& \textit{Bi-CL} &   0.624 &   0.360 &   0.707 &   0.575 &   0.382 &    0.747\\
		& ~~~~\textit{w/gold K} & 0.611  &    0.379 & 0.716 &  0.614  & 0.421  &  0.763 \\
		& ~~~~\textit{- $L_u$} & 0.548 &   0.266 &   0.656 &   0.508 &   0.335 &    0.736\\
		& ~~~~\textit{- $L_s$} & 0.566 &   0.323 &   0.684 &   0.541 &   0.340 &    0.731\\
		& ~~~~\textit{- $L_m$} & 0.596 & 0.345 & 0.697 & 0.542 & 0.341 & 0.731 \\
		& ~~~~\textit{- $L_k$} & 0.612  &  0.282  & 0.643   & 0.133   & 0.100   & 0.589  \\
		\midrule
		\parbox[c]{1mm}{\multirow{1}{*}{\rotatebox[origin=c]{90}{Unsup.~~~}}}
		&\textit{Bi-CL}&   0.607 &   0.345 &   0.704 &   0.581 &   0.366 &    0.689\\
		& ~~~~\textit{w/gold K} & 0.608  &    0.374 & 0.714 &  0.609  & 0.420  &  0.763 \\
		& ~~~~\textit{- $L'_u$} & 0.516 & 0.158   & 0.571   & 0.360   & 0.161   & 0.624   \\
		& ~~~~\textit{- $L'_s$} & 0.607 &   0.337 &   0.640 &   0.570 &   0.354 &    0.683\\
		\bottomrule
	\end{tabular*}
	\caption{\label{ablation} Ablation study on different design components of the proposed \textit{Bi-CL} method under both settings.}
\end{table*}

More specifically, under the supervised setting, the proposed \textit{Bi-CL} method constantly outperforms other methods on the Movie Dialogue dataset across all metrics. It also performs the best on the Ubuntu IRC dataset regarding the metric \textit{Shen-F}. This demonstrates the effectiveness of our Bi-level contrastive learning design for conversation disentanglement. We notice that \textit{DialBERT} and \textit{StructBERT} obtain better \textit{NMI} results on Ubuntu IRC than our method. This is because these methods have special designs to model pairwise relations in a more fine-grained manner, by utilizing additional dialogue features such as the time of each utterance in Ubuntu IRC. Our model omits such data-specific features for model generalizability. In \textit{StructBERT}, the ground truth reference dependencies are leveraged for structural characterization, hence we observe the best \textit{ARI} performance. However, our model indeed surpasses the others on \textit{Shen-F}. Although the margin between the result of our model (0.707) and that of \textit{StructBERT w/max-pooling} (0.681) is smaller than the relatively large margin between the results of \textit{Transition}, \textit{DialBERT}, and \textit{StructBERT}, our model's gain is shown to be statistically significant. We conduct a significance test by running our model in the same setting for 10 times and obtain standard deviation of \textit{NMI} (0.00426), \textit{ARI} (0.00286), and \textit{Shen-F} (0.00186). With the significance level of 0.05, our result for \textit{Shen-F} is significantly superior to the most competitive baseline.

Under the unsupervised setting, our model again excels except for \textit{NMI} in Ubuntu IRC. This might be because the model \textit{Zeroshot} has access to more augmented data from the same data source. However, it still performs worse than \textit{Bi-CL} in \textit{ARI} and \textit{Shen-F}. Note that our model outperforms the baselines with a significant margin on the Movie Dialogue dataset. Again, this implies our model's generalizability. The model \textit{Zeroshot} does not have results on the Movie Dialogue dataset. It relies on same source data to train response selection model, but such data is not available. We also put the our model's results with Silhouette algorithm as the K predictor. There is a slight drop in performance, which can be attributed to the lower prediction accuracy presented on Table \ref{cluster}.

A common pattern shared across the above settings is that while the baselines' results are typically much higher in Ubuntu IRC than in Movie Dialogue, \textit{Bi-CL} performs stably across the two datasets. This is consistent with our previous observation that \textit{Bi-CL} is independent of many features in Ubuntu IRC that are heavily utilized but often not available for other data sources. Moreover, the performance gap between the supervised and unsupervised versions of \textit{Bi-CL} is relatively small, suggesting that it also relies less on labels. These demonstrate the potential of the model to be applied widely.

\subsection{More Analysis}
We further carry out ablation studies on various design components and provide more analysis on the prediction of session number $K$.

\subsubsection{Ablation Study}
We conduct ablation studies to investigate how each model component affects its effectiveness. As shown in Table \ref{ablation}, we observe that in the supervised setting, removing $L_k$ leads to the most significant performance drop, with the gaps of $0.442$, $0.282$ and $0.158$ in \textit{NMI}, \textit{ARI} and \textit{Sehn-F} on the Movie Dialogue. This is because it makes predicting \textit{K} degenerate into a random guess. Also, we observe that $L_u$ has the second most impact. For example, it reaches the lowest performance on Ubuntu IRC regarding \textit{NMI} and \textit{ARI}. Removing the other components has a smaller impact and the model can still generate reasonable result. 

In unsupervised setting, removing $L'_u$ undermines the model significantly on \textit{ARI} since it removes pairwise contrastive learning on the utterance-level that helps to model local relations. Removing $L'_s$ tends to have a milder impact, but it still undermines the results to a certain extent. The above results imply that the utterance-level loss captures local pairwise relations well and the session-level loss also has positive contribution to learning cluster-friendly utterance representations.

\begin{table}[!htp]
	\centering
	\small
	\setlength{\tabcolsep}{3pt}
	\renewcommand*{\arraystretch}{1.3}
	\begin{tabular*}{0.45\textwidth}{lcc|cc}
		\toprule
		& \mc{2}{c}{Ubuntu IRC} & \mc{2}{c}{Movie Dialogue}\\
		\cmidrule{2-3} \cmidrule{4-5}
		& {$~~~ACC~~$} & {$~~MAE~~~$} 
		& {$~~ACC~~$} & {$~~MAE$} \\
		\midrule
		Supervised&  0.272  &    1.389 &  0.682  & 0.330 \\
		Silhouette& 0.166 & 2.085 & 0.251 & 1.074 \\
		Elbow& 0.203 & 1.731 & 0.227 & 1.195 \\
		\bottomrule
	\end{tabular*}
	\caption{\label{cluster} Accuracy (\textit{ACC}) and Mean Absolute Error (\textit{MAE}) of the predictions given by the \textit{K} predictors.}
	\vspace{-0.2cm}
\end{table}

\subsubsection{Prediction of \textit{K}}
Predicting the session number \textit{K} is crucial for our model since it directly affects the clustering results. We hence replace the predicted \textit{K} with the ground truth \textit{K} in training and inference, resulting in a moderate performance boost (\textit{w/gold K}) in both settings as shown in Table \ref{cluster}. We also observe that
the performance gaps between model using predicted \textit{K}  and ground truth \textit{K}. This show that the model with the predicted \textit{K} can still generate relatively satisfactory results and the performance of \textit{K} prediction is relatively good. We show the \textit{ACC} and \textit{MAE} of predicted \textit{K} in Table \ref{cluster}. It indicates that the supervised predictor works better which is reasonable, and the unsupervised methods such as Silhoutte and Elbow perform similarly. This might be because both of them only work on utterance features. Introducing other side information from conversation might further boost the performance.

Another observation is that \textit{NMI} on Ubuntu IRC has a decrease when gold \textit{K} value is adopted in supervised setting. While it is counter-intuitive, it may actually be caused by large number of sessions that contains only one utterance in this dataset. 

\section{Conclusion}
We studied disentanglement on multi-party conversations and proposed a general model that works in both supervised and unsupervised learning settings. It is trained with a Bi-Level contrastive learning mechanism to bring utterances in the same session closer and encourage utterances to flock around their session centers. At the same time, we aim to pull utterances from different sessions further apart by contrasting each utterance with negative samples. The obtained representations naturally fit to the clustering scheme for session predictions. Consequently, K means is used during inference to predict the sessions.
Our model is evaluated on the largest benchmark dataset Ubuntu IRC and the latest benchmark dataset Movie Dialogue. Experimental results show new SOTA performance results and advancements compared to previous works. Additionally, the stability of our model across different datasets, as well as different training schemes with or without session labels, shows its potential to be applied in a general setting, 

\section{Limitations}
Our work has the following limitations. Firstly, although bidirectional LSTM is more light-weight and obtains reasonable performance for our task, an easy extension is to explore how pre-trained language models such as BERT would further affect the performance \cite{liao2021dialogue}. Secondly, the prediction of session number \textit{K} is only based on conversation utterances. More advanced session number estimation model would be devised to capture more side information for more accurate \textit{K} prediction. An alternative approach is to adopt different clustering algorithm such as CISIR \cite{jiang2018} that does not require the prediction of cluster number but instead has a universal, empirically determined threshold that controls the cluster size. 
Last but not least, our model has not been applied to dialogues of length longer than 50, and we have not verified its effectiveness of modeling longer dependency. This entails our future effort to adapt our model to a more general setting with longer conversation, more threads, and more complicated dialogue structures.

\section*{Acknowledgments}
This research was supported by the Singapore Ministry of Education
(MOE) Academic Research Fund (AcRF) Tier 1 grant.

\bibliography{anthology,custom}
\bibliographystyle{acl_natbib}

\newpage
\appendix
\numberwithin{equation}{section}
\section{Appendix}
\label{sec:appendix}
Prototypical Contrastive Learning was originally introduced in \cite{li2020prototypical} to learn image representations. An Expectation-Maximization framework is constructed, where the E step estimates the distribution of the prototypes via K-Means clustering and the M step maximizes the likelihood of the network parameters. 
Similarly, consider a dialogue with $n$ utterances $U = \langle u_1, ..., u_n\rangle $ that are embedded as $\langle \textbf{v}_1, ..., \textbf{v}_n\rangle $. Denote the embedding network parameters as $\theta$, the objective is to find the optimal parameters $\theta^*$ that maximizes the log likelihood of the utterance representations:
\begin{align*}
	\theta^* = \argmax_{\theta} \sum_{i=1}^{|U|} \log p(\textbf{v}_i| \theta).
\end{align*}

Define the set of prototypes in the dialogue as $C = \{c_j\}_{j=1}^{m}$, which are the centroids of the clusters generated by the K-Means algorithm applied on the utterance embeddings. The likelihood for utterance $u_i$ can be written as the summation of the joint probability for $u_i$ being observed and belong to each prototype $c_j$. Hence, we have:
\begin{align*}
	\theta^* &= \argmax_{\theta} \sum_{i=1}^{|U|} \log \sum_{\textbf{c}_j \in C} p(\textbf{v}_i, \textbf{c}_j|\theta)\\
	&= \sum_{i=1}^{|U|} \log \sum_{\textbf{c}_j \in C} Q(\textbf{c}_j) \frac{p(\textbf{v}_i, \textbf{c}_j| \theta)}{Q(\textbf{c}_j)}\\
	&\geq \sum_{i=1}^{|U|} \sum_{\textbf{c}_j \in C} Q(\textbf{c}_j)\log \frac{p(\textbf{v}_i, \textbf{c}_j|\theta)}{Q(\textbf{c}_j)},
\end{align*}
where Jensen's inequality is applied and we have $Q(\textbf{c}_j) = p(\textbf{c}_j|\textbf{v}_i,\theta)$. By ignoring the constant $-\sum_{i=1}^{n} \sum_{\textbf{c}_j \in C}Q(\textbf{c}_j)\log Q(\textbf{c}_j)$, the transformed objective becomes to maximize:
\begin{align}
	L = \sum_{i=1}^{|U|} \sum_{\textbf{c}_j \in C} Q(\textbf{c}_j)\log p(\textbf{v}_i, \textbf{c}_j| \theta).
	\label{object}
\end{align}

\subsection{E step}
In this step, we estimate $Q(\textbf{c}_j) =p(\textbf{c}_j|\textbf{v}_i, \theta)$, which is the likelihood for $u_i$ to be allocated to the cluster with $c_j$ as the centroid. We model $p(\textbf{c}_j|\textbf{v}_i, \theta)$ as $\mathds{1}(u_i \in c_j)$, which is 1 if $u_i$ is allocated to the cluster corresponding to $c_j$ by the K-means algorithm, and 0 otherwise.

\subsection{M step}
In this step, we model Equation \ref{object} and derive the maximization objective. Note that:
\begin{align*}
	p(\textbf{v}_i, \textbf{c}_j| \theta) &= p(\textbf{v}_i|\textbf{c}_j, \theta)p(\textbf{c}_j|\theta)\\
	&= \frac{1}{m}\cdot p(\textbf{v}_i|\textbf{c}_j,\theta),
\end{align*}
since we assume any unseen utterance has an equal probability to belong to any session ($p(\textbf{c}_j|\theta) = \frac{1}{m}$).

Additionally, we assume that the distribution of an utterance $u_i$ around each prototype $c_j$ is an isotropic Gaussian distribution. Therefore, we have:
\begin{align*}
	p(\textbf{v}_i|\textbf{c}_j, \theta) = \frac{\exp{-\frac{(\textbf{v}_i-\textbf{c}_j)^2}{2\sigma_j^2}}}{\sum_{l=1}^{m}\exp{-\frac{(\textbf{v}_i-\textbf{c}_l)^2}{2\sigma_l^2}}} ~.
\end{align*}

We apply $\mathit{l}_2$ normalization to $\textbf{v}_i$ and $\textbf{c}_l$, so that $(\textbf{v}_i - \textbf{c}_l)^2 = 2 - 2\textbf{v}_i\cdot \textbf{c}_l$. As a result, we have:
\begin{align*}
	& \sum_{i=1}^{|U|} \sum_{\textbf{c}_j \in C} Q(\textbf{c}_j)\log p(\textbf{v}_i, \textbf{c}_j|\theta)\\
	& = \sum_{i=1}^{|U|} \sum_{c_j \in C} \mathds{1}(u_i \in c_j)\log \frac{1}{m}p(\textbf{v}_i|\textbf{c}_j, \theta)\\
	& = \sum_{i=1}^{|U|} \log \frac{1}{m}\frac{\exp{(\textbf{v}_i\cdot \textbf{c}_i/\phi_i)}}{\sum_{l=1}^{m}\exp{(\textbf{v}_i\cdot\textbf{c}_l/\phi_l)}},
\end{align*}
where $\phi_l$ indicates the concentration level of the utterance embedding around $c_l$. Here, $\phi_l$ is set as a constant across different prototypes since there are limited number of utterances in the dialogue.

To further enable the learning of contrastive features on different granularity, we cluster the utterance embeddings $M$ times with cluster number ranging from 1 to $M$ and update the network parameters with prototypes that encodes hierarchical structure. Consequently, we can write the optimal parameter as:
\begin{align*}
	\theta^*=\argmin_{\theta} -\frac{1}{M}\sum_{i=1}^{|U|}\sum_{m=1}^M\frac{\exp{(\textbf{v}_i\cdot \textbf{c}_i/\tau'_2)}}{\sum_{l=1}^{m}\exp{(\textbf{v}_i\cdot\textbf{c}_l/\tau'_2)}},
\end{align*}
where $\tau'_2$ is a small constant.


\end{document}